\documentclass[letterpaper, 10 pt, conference]{ieeeconf}  

\usepackage[T1]{fontenc}
\usepackage{pifont}
\usepackage{hyperref}
\usepackage{graphicx}
\usepackage{xspace}
\usepackage{listings}
\usepackage{tcolorbox}
\tcbuselibrary{listingsutf8}
\usepackage{color}
\usepackage[dvipsnames]{xcolor}
\usepackage{tabularx}
\usepackage{booktabs}
\usepackage{multirow}
\usepackage{cite}
\usepackage{algorithm}
\usepackage{algorithmic}
\usepackage{amsmath}
\usepackage{amsfonts}
\usepackage{svg}
\usepackage{stfloats}

\newtcblisting{coloredlisting}{
    listing only,
    colframe=gray!60,
    colback=white,
    boxrule=0.5pt,
    coltitle=black,
    fonttitle=\small\itshape,
    listing options={
        basicstyle=\ttfamily\scriptsize,
        breaklines=true,
        language=Python,
        escapeinside={(*@}{@*)},
        commentstyle=\color{black!40!gray},
    },
    width=\linewidth,
}

\definecolor{baseline_red}{HTML}{E74C3C}
\definecolor{clutter_blue}{HTML}{4C72B0}
\definecolor{routeblue}{RGB}{31,119,180}
\definecolor{routegreen}{RGB}{44,160,44}
\definecolor{routeorange}{RGB}{227,119,37}
\definecolor{routemagenta}{RGB}{214,39,140}

\IEEEoverridecommandlockouts                              

\title{\LARGE \bf
Moving Through Clutter: Scaling Data Collection and Benchmarking for 3D Scene-Aware Humanoid Locomotion via Virtual Reality 
}

\author{Beichen Wang, Yuanjie Lu, Linji Wang, Liuchuan Yu, and Xuesu Xiao
\thanks{All authors are with George Mason University, 4400 University Dr, Fairfax, VA 22030, USA {\tt\small \{bwang25, ylu22, lwang44, lyu20, xiao\}@gmu.edu}}%
\thanks{This work has taken place in the RobotiXX Laboratory at George Mason University. RobotiXX research is supported by National Science Foundation (NSF, 2350352), Army Research Office (ARO, W911NF2320004, W911NF2420027, W911NF2520011), Air Force Research Laboratory (AFRL), US Air Forces Central (AFCENT), Google DeepMind (GDM), Clearpath Robotics, Raytheon Technologies (RTX), Tangenta, Mason Innovation Exchange (MIX), and Walmart.}%
}

\begin{document}

\maketitle

\begin{abstract}


Recent advances in humanoid locomotion have enabled dynamic behaviors such as dancing, martial arts, and parkour, yet these capabilities are predominantly demonstrated in open, flat, and obstacle-free settings.
In contrast, real-world environments—homes, offices, and public spaces—are densely cluttered, three-dimensional, and geometrically constrained, requiring scene-aware whole-body coordination, precise balance control, and reasoning over spatial constraints imposed by furniture and household objects. 
However, humanoid locomotion in cluttered 3D environments remains underexplored, and no public dataset systematically couples full-body human locomotion with the scene geometry that shapes it.
To address this gap, we present Moving Through Clutter (MTC), an open-source Virtual Reality (VR)-based data collection and evaluation framework for scene-aware humanoid locomotion in cluttered environments.
Our system procedurally generates scenes with controllable clutter levels and captures embodiment-consistent, whole-body human motion through immersive VR navigation, which is then automatically retargeted to a humanoid robot model. We further introduce benchmarks that quantify environment clutter level and locomotion performance, including stability and collision safety. Using this framework, we compile a dataset of 348 trajectories across 145 diverse 3D cluttered scenes. The dataset provides a foundation for studying geometry-induced adaptation in humanoid locomotion and developing scene-aware planning and control methods.

\end{abstract}

\section{Introduction}
\label{sec:intro}

Humanoid locomotion on flat terrain has witnessed rapid progress in recent years, with learning-based controllers achieving highly dynamic behaviors such as agile running, recovery from perturbations, and acrobatic motion synthesis \cite{kim2024acrobatic,wang2026omnixtreme}. A central driver of this progress is the availability of large-scale human motion datasets. By leveraging motion priors derived from resources such as AMASS dataset \cite{mahmood2019amass}, reinforcement learning policies can acquire diverse and natural locomotion strategies that would be difficult to engineer manually. The scale and diversity of motion data have emerged as a key factor shaping the capabilities of modern humanoid locomotion systems.

Despite these advances, most existing methods are developed and evaluated in open, obstacle-free settings. Real-world deployment, however, requires humanoid robots to operate within cluttered environments where locomotion is constrained by surrounding three-dimensional (3D) geometry, like in homes or offices. Locomotion in such settings demands continuous whole-body postural adaptation coupled to environmental structure, including lateral clearance management, height-aware motion, and acyclic and asymmetric limb adjustment to avoid collisions. These requirements fundamentally differ from flat-ground locomotion and remain largely underexplored from a data-centric perspective.

Although human motion datasets have proven essential for humanoid locomotion learning, no publicly available resource currently provides scene-aware, embodiment-consistent humanoid locomotion trajectories that are explicitly aligned with 3D cluttered environments. The primary bottleneck lies in scalable data acquisition. Traditional motion capture pipelines rely on open studios without occlusions and therefore fail to encode interactions with spatial constraints. Constructing diverse physical environments, e.g., with large furniture and household objects, for data collection is expensive and difficult to reproduce at scale. 
Meanwhile, recent Virtual Reality (VR)-based teleoperation systems demonstrate that immersive interfaces can bridge human motion and robot embodiment\cite{ze2025twist2,luo2025sonic}, yet these efforts are primarily designed for real-time control rather than systematic dataset construction.

\begin{figure}[t]
\centering
\includegraphics[width=\columnwidth]{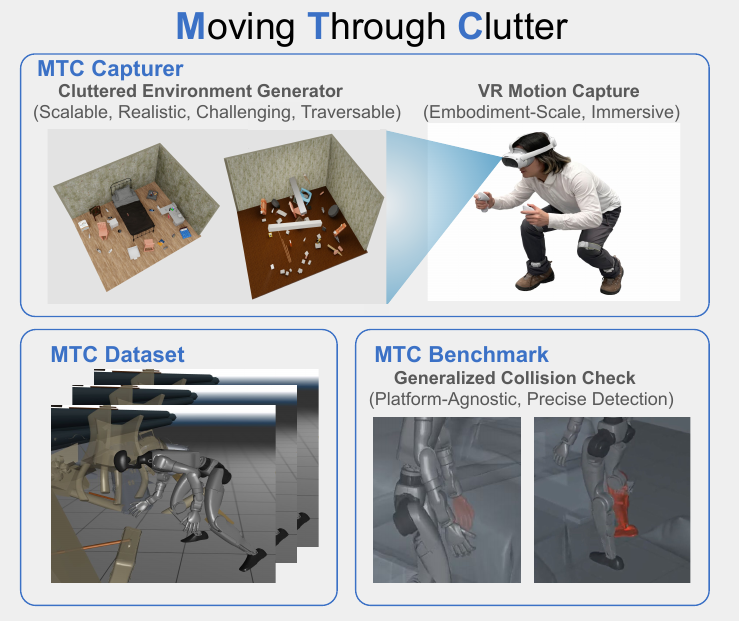}
\vspace{-16pt}
\caption{
We introduce MTC, a dataset and benchmark for humanoid locomotion in cluttered 3D environments. We first procedurally generate cluttered simulation scenes with diverse geometric constraints, then collect immersive, embodiment-scaled locomotion trajectories using VR-based full-body tracking within these environments.
The resulting dataset enables benchmarking of scene-aware humanoid locomotion behaviors.
}
\label{fig:mtc_overview}
\vspace{-16pt}
\end{figure}

Motivated by these limitations, we introduce an open-source framework, Moving Through Clutter (MTC), for collecting and evaluating scene-aware humanoid locomotion data. Our approach combines procedural cluttered environment generation with embodiment-scaled virtual reality capture. By scaling the human operator to match the physical proportions of the target humanoid during data acquisition, the recorded trajectories become geometrically consistent with robot-scale locomotion, effectively transforming human-scale demonstrations into embodiment-aligned motion data. The entire pipeline operates in virtual environments without requiring physical scene construction, enabling scalable and reproducible dataset generation under controlled clutter level and geometric difficulty. To systematically validate the collected trajectories, we further introduce a quantitative evaluation benchmark that measures locomotion difficulty and collision safety under geometric constraints, providing a standardized protocol for assessing scene-aware humanoid locomotion.
The contributions of our MTC framework are summarized as follows:

\begin{itemize}
\item MTC Capturer: An embodiment-scaled VR capture paradigm that produces geometrically consistent locomotion trajectories aligned with humanoid embodiment and procedurally generated 3D scenes with continuous clutteredness parameterization;
\item MTC Dataset: An open-source dataset that jointly provides whole-body locomotion trajectories and corresponding 3D scene configurations; and
\item MTC Benchmark: A dual evaluation benchmark that quantitatively measures locomotion difficulty and collision safety in geometrically constrained environments.
\end{itemize}

\begin{figure*}[t]
    \centering
    \includegraphics[width=\textwidth]{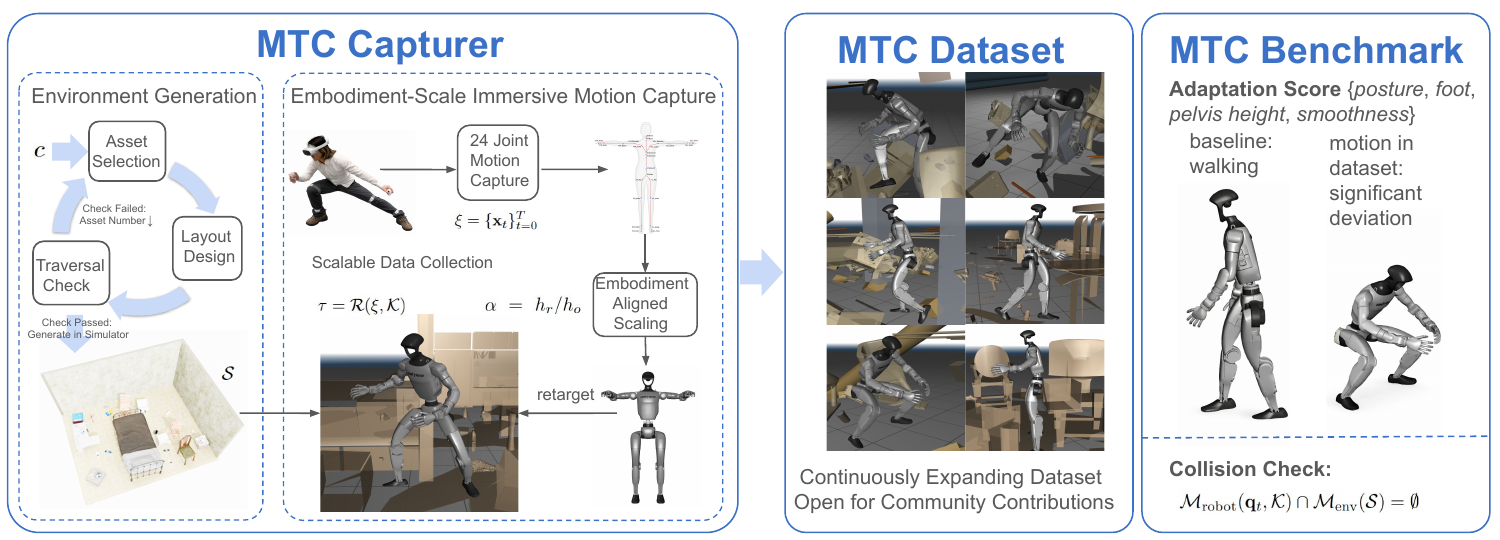}
    \vspace{-10pt}
    \caption{
    System overview of \textbf{MTC}.  \textbf{MTC Capturer}: immersive VR-based system for collecting embodiment-scaled human locomotion trajectories in cluttered environments.
    \textbf{MTC Dataset}: a large-scale collection of locomotion trajectories recorded in procedurally generated cluttered scenes.
    \textbf{MTC Benchmark}: evaluation framework for measuring scene-aware locomotion behaviors relative to normal walking.
    }
    \label{fig:mtc_system_overview}
    \vspace{-12pt}
\end{figure*}

\section{Related Works}
\label{sec:related}
We review related work in learning humanoid locomotion, human motion datasets and VR teleoperation, and procedural simulation environment generation. 

\subsection{Learning-Based Humanoid Locomotion}

Reinforcement learning has enabled significant progress in humanoid locomotion. Radosavovic et al. \cite{radosavovic2024humanoid} and Haarnoja et al. \cite{haarnoja2024soccer} demonstrated robust sim-to-real transfer and agile dynamic behaviors, establishing that learning-based controllers can achieve stable and versatile locomotion. Human motion priors further accelerate this progress: DeepMimic \cite{peng2018deepmimic} and Adversarial Motion Priors (AMP) \cite{peng2021amp} introduced example-guided and style-based rewards derived from human demonstrations, while ExBody \cite{cheng2024exbody}, ExBody2 \cite{ji2024exbody2}, and SONIC \cite{luo2025sonic} scaled whole-body tracking with large motion datasets.

Despite these advances, existing policies are predominantly trained and evaluated on flat or sparsely structured terrain, where the primary challenge is balance and contact stability. Even terrain-variant settings typically consist of engineered primitives such as slopes or isolated obstacles. In contrast, locomotion in cluttered 3D indoor environments requires continuous modulation of torso orientation, lateral clearance, and full-body posture to avoid multi-height collisions. Current approaches achieve dynamic locomotion yet rarely condition control on detailed scene geometry, leaving geometry-conditioned traversal strategies largely unexplored.

While Xue et al. \cite{xue2025humanoidpf} make an important step toward cluttered indoor traversal by introducing HumanoidPF as an informative representation and training traversal behaviors purely with RL, their formulation is primarily optimized for collision avoidance learnable via strong, hand-designed directional guidance---HumanoidPF is explicitly queried at body parts as policy observation and also shapes dense reward signals. As a result, the learned behaviors are naturally biased toward a compact set of goal-directed traversal skills (e.g., hurdle/crouch/squeeze) that can be efficiently acquired under such guidance, rather than toward rich, human-like, and reusable geometry-conditioned locomotion strategies grounded in large-scale demonstrations. This leaves open how to systematically capture and leverage diverse full-body locomotion data paired with the exact 3D scene geometry that induces it, so that policies can be learned not only to be collision-free, but also to be expressive, natural, and broadly reusable across clutter distributions.

\subsection{Human Motion Datasets and VR Teleoperation}

Large-scale motion capture datasets such as AMASS \cite{mahmood2019amass}, BABEL \cite{punnakkal2021babel}, HumanML3D \cite{guo2022humanml3d}, and Motion-X \cite{lin2023motionx} provide diverse human motion sequences and semantic annotations, substantially advancing motion modeling. However, these datasets are recorded in obstacle-free studios and do not encode explicit coupling between motion and environmental geometry, preventing models from learning how locomotion adapts to spatial constraints.

Scene-aware datasets including PROX \cite{hassan2019prox}, SAMP \cite{hassan2021samp}, HUMANISE \cite{wang2022humanise}, and TRUMANS \cite{jiang2024trumans} introduce motion aligned with 3D indoor scenes. Yet they focus primarily on human motion prediction and generation characterized by quasi-static interactions from the computer vision perspective, limiting applicability to humanoid locomotion with different kinematics and actuation constraints, particularly for sustained traversal through narrow or irregular passages.


VR-based teleoperation systems \cite{he2024h2o, he2024omnih2o, cheng2024opentv, fu2024humanplus, ze2025twist, ze2025twist2} and retargeting frameworks \cite{araujo2026gmr, yang2025omniretarget} demonstrate effective human-to-humanoid motion transfer. However, these efforts emphasize real-time control for loco-manipulation rather than systematically constructing reusable locomotion datasets paired with procedurally varied environments and quantitative benchmarking protocols.

\subsection{Procedural Simulation Environment Generation}


Procedural environment generation has primarily supported embodied navigation and visual understanding. ProcTHOR \cite{deitke2022procthor}, Infinigen \cite{raistrick2023infinigen, raistrick2024infinigen_indoors}, Holodeck \cite{yang2024holodeck}, and Habitat \cite{savva2019habitat} generate large-scale interactive indoor or outdoor scenes with semantic coherence and navigational accessibility.

However, humanoid locomotion in cluttered indoor spaces is constrained by full-body volumetric clearance, multi-height collision risks, and kinematic feasibility. Existing procedural generators neither model embodiment-aware traversal constraints nor parameterize scene difficulty in terms of clearance margins or collision risk. As a result, they are not designed to support geometry-conditioned humanoid locomotion learning or benchmarking.

Taken together, prior work has advanced locomotion control, motion modeling, teleoperation, and scene generation largely in isolation. What remains missing is a unified framework that couples procedural clutter generation, immersive embodiment-scaled locomotion data capture scalable by VR, motion retargeting, and quantitative benchmarking for scene-aware humanoid traversal in cluttered 3D environments.

\section{MTC Capturer}
\label{sec:capturer}
We now formalize the 3D scene-aware humanoid locomotion task and describe the MTC Capturer, the core data acquisition module that enables embodiment-scaled motion capture within generated cluttered environments in VR.

\subsection{Problem Formulation}

We consider goal-directed humanoid locomotion in cluttered 3D environments. Let a humanoid robot with kinematic model $\mathcal{K}$ operate in a scene $\mathcal{S}$.  The robot configuration space is denoted by $C$, where a configuration $\mathbf{q} \in C$ encodes all joint configurations of the robot.
A goal-directed humanoid locomotion task is specified by an initial configuration $\mathbf{q}_0 \in C$ and a goal set $\mathcal{G} \subset C$. 
A trajectory is defined as a time-ordered sequence of configurations
\[
\tau = \{\mathbf{q}_t\}_{t=0}^{T}, \qquad \mathbf{q}_t \in C,
\]
which is considered successful if its terminal configuration satisfies
\[
\mathbf{q}_T \in \mathcal{G}.
\]
A traversal trajectory is considered valid only if it satisfies the following necessary conditions throughout execution:

\begin{itemize}
    \item \textbf{Collision safety.}
    The robot must remain collision-free with respect to the scene geometry at all times:
    \begin{equation}
    \mathcal{M}_{\mathrm{robot}}(\mathbf{q}_t, \mathcal{K})
    \cap
    \mathcal{M}_{\mathrm{env}}(\mathcal{S})
    =
    \emptyset,
    \quad
    \forall t \in \{0,\dots,T\}.
    \nonumber
    \end{equation}
    Here $\mathcal{M}_{\mathrm{robot}}(\mathbf{q}_t, \mathcal{K})$ denotes the robot body mesh obtained via forward kinematics and $\mathcal{M}_{\mathrm{env}}(\mathcal{S})$ denotes the environment geometry.

    \item \textbf{Postural stability.}
    The robot must maintain dynamic support throughout execution and must not undergo irrecoverable loss of balance.
\end{itemize}
We denote by $\mathcal{T}(\mathcal{S}, \mathcal{K}, \mathcal{G})$ the set of trajectories that satisfy both goal completion defined by $\mathcal{G}$ and the above physical feasibility conditions defined by $\mathcal{S}$ and $\mathcal{K}$.

\subsection{Data-Driven Solutions with MTC Capturer}
Learning-based approaches require a humanoid robot dataset of embodiment-specific traversal trajectories,
\[
\mathcal{D}^{\mathrm{robot}}
=
\{(\mathcal{S}_i, \mathcal{K}, \mathcal{G}_i, \tau_i)\}_{i=1}^{N},
\quad
\tau_i \in \mathcal{T}(\mathcal{S}_i, \mathcal{K}, \mathcal{G}_i),
\]
where each $\tau_i$ is a physically feasible, goal-directed humanoid locomotion trajectory executed under the kinematic embodiment $\mathcal{K}$ in a cluttered scene $\mathcal{S}_i$.

However, constructing $\mathcal{D}^\textrm{robot}$ directly with real humanoid platforms is impractical. 
First, constructing geometry-conditioned locomotion data requires a diverse distribution of scene geometries $\mathcal{S}$ that impose structured 3D spatial constraints. Physical construction of such environments is costly and difficult to regulate systematically. Second, collecting collision-free and stable humanoid trajectories in such environments presupposes solving the scene-aware locomotion problem under study; moreover, human demonstration from direct teleoperation of highly articulated humanoids under tight geometric constraints is extremely challenging. 

To overcome these limitations, we introduce the \textbf{MTC Capturer}, 
a virtual-reality data collection pipeline that enables scalable scene generation and embodiment-consistent motion acquisition in VR.
The MTC Capturer first procedurally generates geometrically diverse cluttered scenes in simulation to enable scalable environment construction, avoiding the need to set up any physical geometry in the real world.  
After that, it captures embodiment-scaled human reference motions as operators traverse these virtual environments, allowing geometric constraints to induce whole-body adaptation at the target robot's proportions.
Let
\[
\xi = \{\mathbf{x}_t\}_{t=0}^{T}
\]
denote a human reference motion sequence collected. The resulting motion is subsequently mapped to the humanoid configuration space through a retargeting function
\[
\tau = \mathcal{R}(\xi, \mathcal{K}),
\]
yielding robot traversal trajectories consistent with embodiment $\mathcal{K}$.
MTC Capturer's objective is therefore to construct a reproducible embodiment-consistent dataset that couples procedurally generated scene geometries with immersive human reference motions,
\[
\mathcal{D}^{\mathrm{MTC}}
=
\{(\mathcal{S}_i, \mathcal{K}, \mathcal{G}_i, \xi_i)\}_{i=1}^{N},
\]
where each sample consists of a generated scene $\mathcal{S}_i$ and an embodiment-scaled human reference motion $\xi_i$. 
Corresponding robot trajectories are obtained via retargeting.

The two components of MTC Capturer are detailed in the following subsections. The first component addresses scalable scene construction, while the second enforces embodiment-consistent geometry during motion acquisition.

\subsection{Procedural Environment Generation}
A locomotion-relevant cluttered scene $\mathcal{S}$ must satisfy three essential requirements:

\begin{itemize}
    \item It must preserve the semantic characteristics of a real-world scenario. Scenes should reflect recognizable room types (e.g., bedroom, living room, and kitchen) with functionally meaningful object arrangements rather than arbitrary object scattering;
    \item It must exhibit sufficient geometric clutter such that traversal cannot be accomplished by trivial straight-line walking. Instead, the environment should induce constraint-driven whole-body adaptation, such as side-stepping through narrow passages, torso reorientation, crouching under height-constrained obstacles, or clearance-aware stepping; and
    \item It must guarantee at least one feasible traversal path from a start to a goal for the target humanoid.
\end{itemize}
MTC Capturer addresses these three requirements through a structured procedural pipeline described below.

\subsubsection{Geometric Regimes}
\begin{figure}[t]
\centering
\includegraphics[width=0.98\columnwidth]{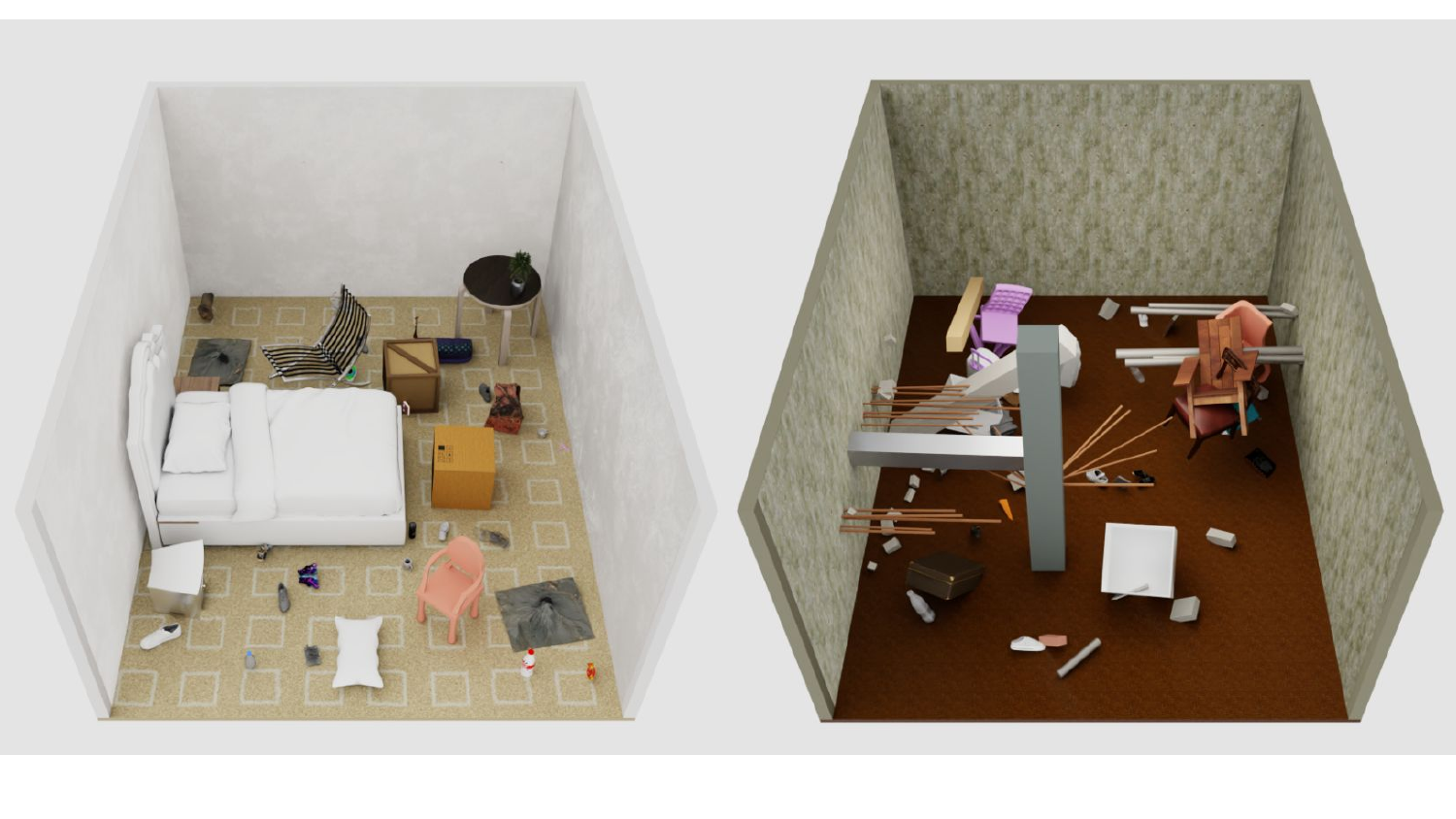}
\vspace{-16pt}
\caption{Examples of cluttered environments from two geometric regimes in MTC: structured domestic layouts (left) and debris-style layouts (right).
}
\label{fig:mtc_regimes}
\vspace{-16pt}
\end{figure}

Each scene is generated under one of two geometric regimes. 
The first structured domestic regime models semantically organized indoor layouts dominated by furniture-induced lateral confinement and corridor-like free space. 
The second debris-style regime introduces irregular obstacle configurations that create both planar entanglement and vertical clearance restrictions. 
In addition to dense and non-axis-aligned ground-level clutter, this regime incorporates overhead beams and structural elements that constrain height clearance, jointly inducing behaviors such as crouching, ducking, or crawling. Representative examples of the two regimes are shown in Fig.~\ref{fig:mtc_regimes}.

\subsubsection{Semantic Layout Structure}

Each scene begins by sampling a geometric regime and room type. To preserve structural coherence while enabling controllable variation, assets are organized into functional tiers and placed in a hierarchical order:
\begin{itemize}

    \item \textbf{Anchor Element Layer:} 
    A dominant structural or functional component that defines the primary spatial organization of the scene. In structured domestic layouts, this corresponds to a functional core object (e.g., bed, sofa, and dining table), around which the room is organized. In the debris regime, the anchor role is fulfilled by major structural elements (e.g., load-bearing pillars or primary beam assemblies) that determine the global obstruction pattern. The anchor element is instantiated first according to regime-specific placement priors, ensuring plausible spatial structure.

    \item \textbf{Supporting Large Elements Layer:} 
    Large objects or structures that establish the dominant topology of the environment. In domestic layouts, these include wall-aligned furniture such as wardrobes, cabinets, or bookshelves. In the debris regime, these may include secondary pillars or extended structural members. Placement is performed via probabilistic sampling conditioned on room type and size tier, followed by rejection checks to avoid overlap.

    \item \textbf{Small Clutter Layer:} 
    Density-controlling objects that perturb navigable free space at a finer scale. These include freestanding items (e.g., chairs) as well as scattered small obstacles. Small clutter is injected after the primary layout has been established, allowing density to vary without altering global structure.

    \item \textbf{Vertical Obstruction Layer (debris regime only):} 
    Additional structural members (e.g., overhead beams, rebars, and I-shaped steel) are attached to pillars or walls to introduce height-clearance constraints. These elements create coupled planar and vertical restrictions, inducing crouching, ducking, or crawling behaviors beyond lateral navigation.

\end{itemize}
During generation, anchor elements are instantiated first, followed by supporting large elements, and finally small clutter and regime-specific obstructions. 
This hierarchical layout strategy ensures that global spatial structure is determined before local density perturbations are introduced, preserving semantic plausibility while progressively shaping the navigable topology.

Two diagonally opposite spawn zones, denoted as $\mathcal{Z}_{\mathrm{start}}$ and $\mathcal{Z}_{\mathrm{goal}}$, are reserved as obstacle-free regions prior to placement, defining an origin--destination traversal pair.

\subsubsection{Clutterness-Driven Density Control}
To provide continuous and interpretable control over scene density, we introduce a scalar parameter $c \in [0,1]$ representing the target floor-occupancy ratio.
Rather than manually specifying discrete object counts, the number of items injected into each placement layer is computed from the room area and the typical footprint size of assets in that layer.

Let $A$ denote the room floor area. We allocate the clutterness budget only to non-anchor placement layers. Specifically, for each density-controlled layer 
$i \in \{\texttt{Supporting Large Elements}, \texttt{Small Clutter}\}$,
the target number of items is determined by
\[
n_i = \left\lfloor \frac{c \, A \, w_i}{\bar{a}_i} \right\rfloor,\quad \sum_i w_i = 1. 
\]
Here, $w_i$ is the weight that distributes the occupancy budget between large structural elements and small clutter, while the anchor element is instantiated independently of $w_i$ and $c$. $\bar{a}_i$ denotes the mean ground-plane footprint area of all assets that can be assigned for that layer. The footprint $\bar{a}_i$ is computed at runtime from each asset’s axis-aligned bounding box on the floor plane.
This formulation ensures that the number of items scales proportionally with room size, and that larger assets naturally result in fewer instances under the same clutterness parameter $c$ due to their increased footprint.

While the clutterness parameter $c$ controls obstacle density in expectation, high-density configurations may inadvertently eliminate feasible traversal paths for the target morphology. We therefore perform an explicit navigability verification step after object placement.

\subsubsection{Morphology-Aware Navigability Verification}

After object placement, scene feasibility is validated through a 2D grid-based reachability test.
The floor is discretized at a fixed resolution. Before rasterization, each obstacle footprint is inflated by a configurable humanoid clearance radius $K$, derived from the embodiment descriptor $\mathcal{K}$, effectively constructing a morphology-aware configuration-space map.
The occupancy grid is then constructed from these inflated obstacles, and Breadth First Search is performed on the resulting free-space map. The search is initialized from all free cells within $\mathcal{Z}_{\mathrm{start}}$, and the scene is considered traversable if any cell within $\mathcal{Z}_{\mathrm{goal}}$ can be reached. 
If no feasible traversal path exists, the scene is not discarded outright. Instead, we invoke a constraint-preserving annealed resampling procedure to restore connectivity while maintaining structural coherence.

\subsubsection{Constraint-Preserving Annealed Resampling}
If a candidate layout fails the reachability check, the system applies an annealing schedule that progressively removes assets while preserving structural elements.
Small clutter---the most expendable layer---is reduced first via multiplicative decay (20\% per annealing level), as these elements primarily perturb local free space without defining global structure.  If connectivity remains unsatisfied, supporting large elements are subsequently reduced starting from the next annealing level using the same decay rate. Regime-defining anchor elements and major structural furniture are preserved until higher annealing stages, ensuring that the global spatial organization of the scene remains intact as long as the scene is physically navigable. This hierarchical relaxation policy defines a monotonic density schedule and guarantees convergence to a feasible layout given an appropriate scene size within a bounded number of retries while preserving structural coherence.

Following annealed resampling, the realized clutterness level may differ from the originally specified target $c$. We therefore define the realized density
\[
c' = \frac{\sum_{k} a_k}{A},
\]
where $a_k$ is the ground-plane footprint of each non-anchor object in the final layout. While $c$ controls density in expectation during sampling, $c'$ reflects the actual floor occupancy ratio after reachability-constrained annealing. 

Collectively, hierarchical layout construction, morphology-aware verification, and constraint-preserving annealing yield semantically structured, density-controlled environments whose navigability is explicitly conditioned on the target humanoid embodiment.

\subsection{Embodiment-Scaled Immersive Motion Capture}

Standard motion-capture datasets record human motion at natural body proportions. When retargeted to a robot with different stature, geometric clearances experienced during capture may no longer correspond to those encountered by the robot, leading to unintended collisions or infeasible postures.
To mitigate this embodiment mismatch, we enforce embodiment-consistent geometry during data collection.

To match the scale of the humanoid and human operator, let $h_r$ denote the standing height of the target robot and $h_o$ the measured standing height of the human operator. 
We define the embodiment scale factor as $\alpha = h_r / h_o$.
During data collection, the virtual environment is rendered with a uniform scale factor $1/\alpha$, such that spatial clearances experienced by the operator correspond to those encountered by the robot at its embodiment scale.
Joint poses are recorded at human scale and uniformly rescaled by the factor $\alpha$ after capture, yielding an embodiment-consistent motion sequence expressed in the robot’s spatial scale.

To match the skeletal structure of the humanoid and human operator, existing retargeting framework can be employed. For this work, we use General Motion Retargeting framework of Ze et al.~\cite{ze2025twist2} without modification.

\section{MTC Dataset and Benchmark}
\label{sec:dataset}
We introduce the MTC Dataset and Benchmark to support learning approaches and systematic evaluation of scene-aware humanoid locomotion in cluttered environments. 

\subsection{MTC Dataset}
\subsubsection{Dataset Overview}

The MTC dataset consists of 145 procedurally generated scenes $\{\mathcal{S}_k\}$ and 348 associated traversal trajectories $\{\xi_j\}$ collected under the Unitree G1 humanoid embodiment. 
Scenes span both geometric regimes, i.e., domestic and debris-style, and multiple semantic room types, e.g., bedroom, living room, and kitchen. 
While the target clutterness parameter $c$ is defined over $[0,1]$, empirical analysis under the G1 embodiment indicates that realised densities $c'$ in the range $[0.2,\,0.6]$ most frequently correspond to geometrically challenging yet traversable layouts.

Human motion is captured using a PICO 4 Ultra VR system with integrated full-body tracking, providing 24-joint skeletal pose measurements per frame. In total, the current dataset contains approximately 731,000 motion frames across 348 trajectories, corresponding to roughly 2.3 hours of humanoid locomotion data. Individual trajectories range from 431 to 6,939 frames in length, with a mean of 2,101 frames. 
The MTC dataset is actively expanding and the MTC capturer pipeline will be open-sourced to encourage community contributions.

\subsubsection{Scene Density Distribution}

\begin{figure}[t]
  \centering
  \includegraphics[width=\linewidth]{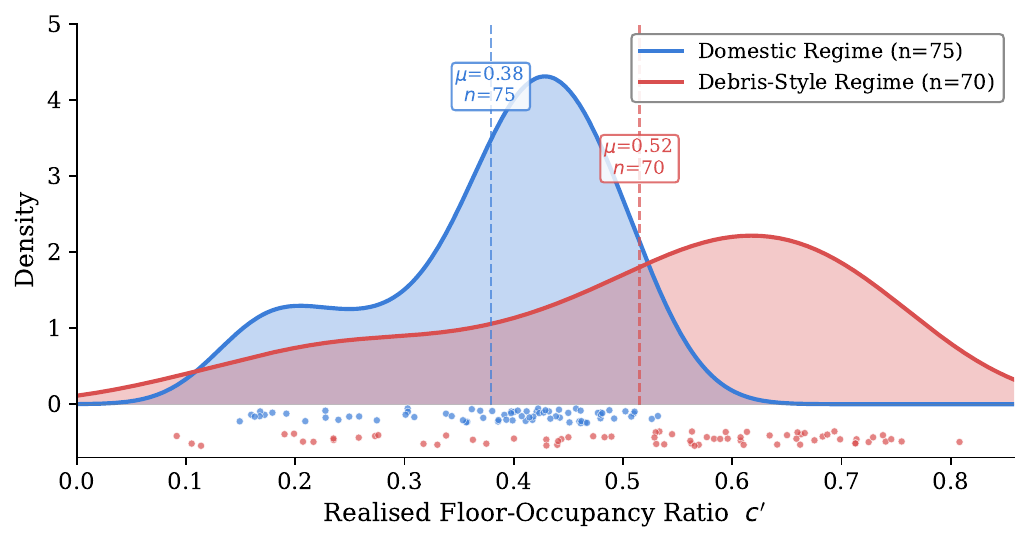}
  \vspace{-16pt}
  \caption{Distribution of realised floor-occupancy ratio $c'$ across 145 generated scenes. Individual scene values are shown as jittered strips below the density curves; dashed lines indicate per-regime means.}
  \label{fig:clutterness_dist}
  \vspace{-16pt}
\end{figure}

To characterize geometric variability across the dataset, we analyze the distribution of realised clutterness levels $c'$ over all generated scenes. Fig.~\ref{fig:clutterness_dist} shows the kernel density estimates of $c'$ for both regimes, computed as the fraction of room floor area occupied by non-anchor object footprints on a 5\,cm rasterization grid.

\subsubsection{Case Study: Goal-Conditioned Route Diversity}

\begin{figure*}[t]
\centering
\includegraphics[width=0.98\textwidth]{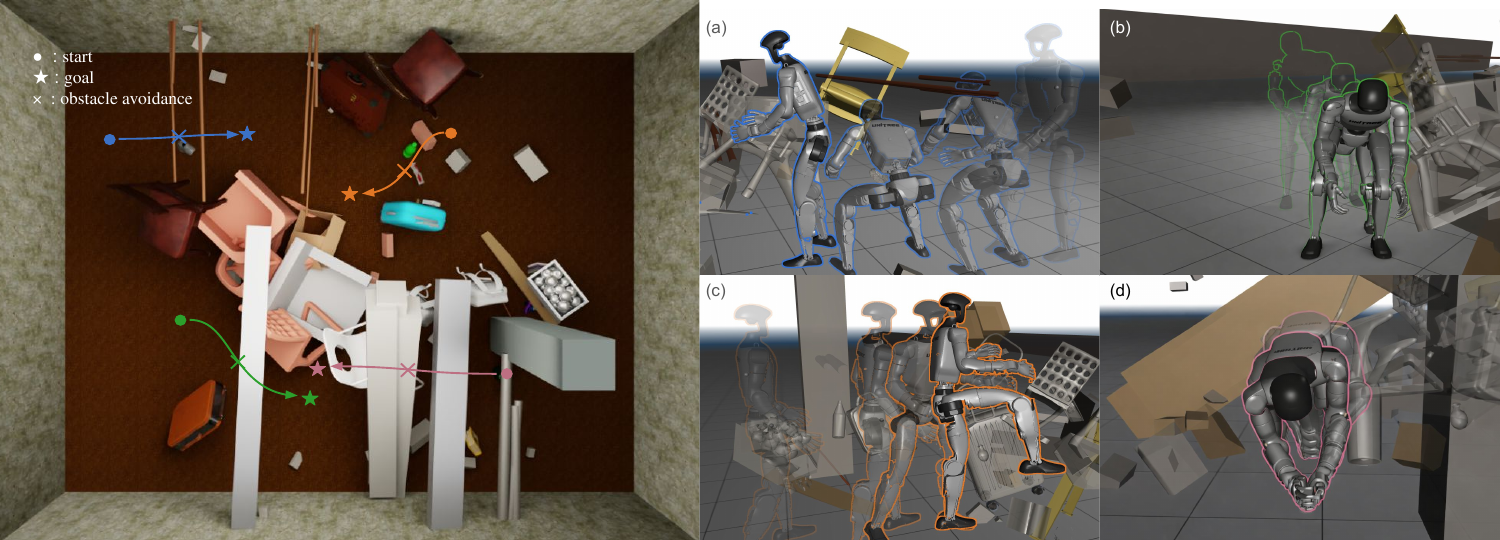}

\caption{
Case study of goal-conditioned route diversity in a representative MTC scene.
The left panel visualizes the scene floorplan together with the ground-plane projections of pelvis trajectories under four goal configurations.
Markers $\bullet$, $\star$, and $\times$ denote start locations, goal positions, and obstacle-avoidance maneuvers, respectively.
Insets (a–d) show representative locomotion behaviors observed along the corresponding routes:
(a) \textcolor{routeblue}{crouched lateral shuffling},
(b) \textcolor{routegreen}{crouched forward shuffling},
(c) \textcolor{routeorange}{high-knee lateral step-over},
and (d) \textcolor{routemagenta}{prone crawling}.
}

\label{fig:route_diversity}

\vspace{-6pt}
\end{figure*}

To illustrate the richness of geometry-induced behavior, we analyze a representative scene under four goal configurations.
Fig.~\ref{fig:route_diversity} presents the scene floor plan together with the ground-plane projections of pelvis trajectories, along with representative snapshots of the corresponding obstacle-avoidance behaviors.
Markers $\bullet$, $\star$, and $\times$ denote start locations, goal positions, and obstacle avoidance maneuvers, respectively.

Although all motions occur within the same environment, 
different goal placements induce distinct traversal routes that expose the agent to different local geometric constraints—such as narrow passages, low-clearance structures, or densely cluttered regions.
This example illustrates that behavioral diversity in MTC arises not only from varying scene layouts, but also from goal-conditioned routing within a single environment.

\subsection{MTC Benchmark}

\subsubsection{Motion Adaptation Score}

Locomotion under geometric constraints induces coordinated deviations from nominal flat-ground walking across multiple kinematic dimensions. 
To quantify geometry-induced whole-body adaptation, we evaluate trajectory statistics within four complementary subspaces:

\begin{itemize}
    \item \emph{Posture}: pelvis-relative joint positions and velocities, capturing configuration-level whole-body adjustments;
    \item \emph{Vertical motion}: pelvis height, vertical velocity, and vertical acceleration, reflecting height-clearance adaptation;
    \item \emph{Foot interaction}: foot heights and vertical velocities, characterizing stepping and obstacle negotiation; and
    \item \emph{Smoothness}: third-order positional differences (jerk) across joints, measuring dynamic modulation under constraint.
\end{itemize}
For each subspace, frame-wise feature vectors are extracted and summarized by their empirical mean and covariance. 
A reference distribution is estimated from a corpus of temporally aligned, speed-normalized flat-ground walking trajectories and approximated by a multivariate Gaussian. 
Given a test trajectory, we compute the Fréchet distance between its empirical feature distribution and the baseline distribution within the same subspace. 
The Fréchet distance captures both mean displacement and covariance shift, enabling joint evaluation of first- and second-order kinematic deviations.

Formally, let $(\mu_r, \Sigma_r)$ and $(\mu_t, \Sigma_t)$ denote the reference and test feature statistics, respectively. 
The squared Fréchet distance is computed as
\begin{equation}
d^2 = \|\mu_r - \mu_t\|^2
+ \mathrm{Tr}\!\left(
\Sigma_r + \Sigma_t
- 2(\Sigma_r^{1/2}\Sigma_t\Sigma_r^{1/2})^{1/2}
\right).
\nonumber
\end{equation}
Subspace distances are normalized and aggregated via uniform weighting to produce a scalar adaptation score. 
Higher values indicate greater deviation from nominal flat-ground locomotion. 
The score is intended as a relative deviation measure rather than an absolute proxy for task difficulty.

\subsubsection{Collision Safety Assessment}

Physical feasibility within cluttered 3D geometry is treated as an explicit evaluation criterion in the MTC Benchmark. 
Given an evaluated trajectory $\{\mathbf{q}_t\}_{t=1}^{T}$ and scene mesh $\mathcal{M}$, collision safety is assessed via signed distance queries against the original (non-convex) scene geometry.

For each frame $t$, forward kinematics maps the joint configuration $\mathbf{q}_t$ to world-frame link poses. 
Surface sample points are generated on each link and queried against a precomputed signed distance field of $\mathcal{M}$. 
Let $s_{t,i}$ denote the signed distance of sample point $i$ at frame $t$, where negative values indicate penetration into the scene geometry.
The frame-wise penetration depth is then defined as
\[
d_t = \max\bigl(0,\; -\min_i s_{t,i}\bigr),
\]
so that $d_t = 0$ indicates a collision-free configuration and $d_t > 0$ measures the maximum instantaneous penetration depth at frame $t$.
Based on $\{d_t\}_{t=1}^{T}$, the benchmark reports four quantitative safety metrics:
\begin{align}
R_{\text{col}} &= \frac{1}{T}\sum_{t=1}^{T} \mathbf{1}[d_t > 0], \nonumber\\
d_{\max} &= \max_{t} d_t, \nonumber\\
\bar{d}_{\text{cond}} &= 
\frac{\sum_{t} d_t \mathbf{1}[d_t > 0]}
     {\max\!\left(1,\sum_{t} \mathbf{1}[d_t > 0]\right)}, \nonumber\\
I_{\text{pd}} &= \frac{1}{T}\sum_{t=1}^{T} d_t.\nonumber
\end{align}
Here, $R_{\text{col}}$ measures collision frequency over the trajectory, 
$d_{\max}$ captures the worst-case geometric violation, 
$\bar{d}_{\text{cond}}$ quantifies average penetration severity conditioned on collision events (defined as zero when no collision occurs), and 
$I_{\text{pd}}$ reflects time-normalized penetration magnitude across the full traversal. 
Together, these metrics characterize both discrete collision occurrence and continuous penetration severity under full scene geometry. The evaluation code will be released as part of the MTC framework to ensure reproducibility.

\subsubsection{Dataset-Level Benchmark Statistics}

\begin{figure}[t]
\centering
\includegraphics[width=\columnwidth]{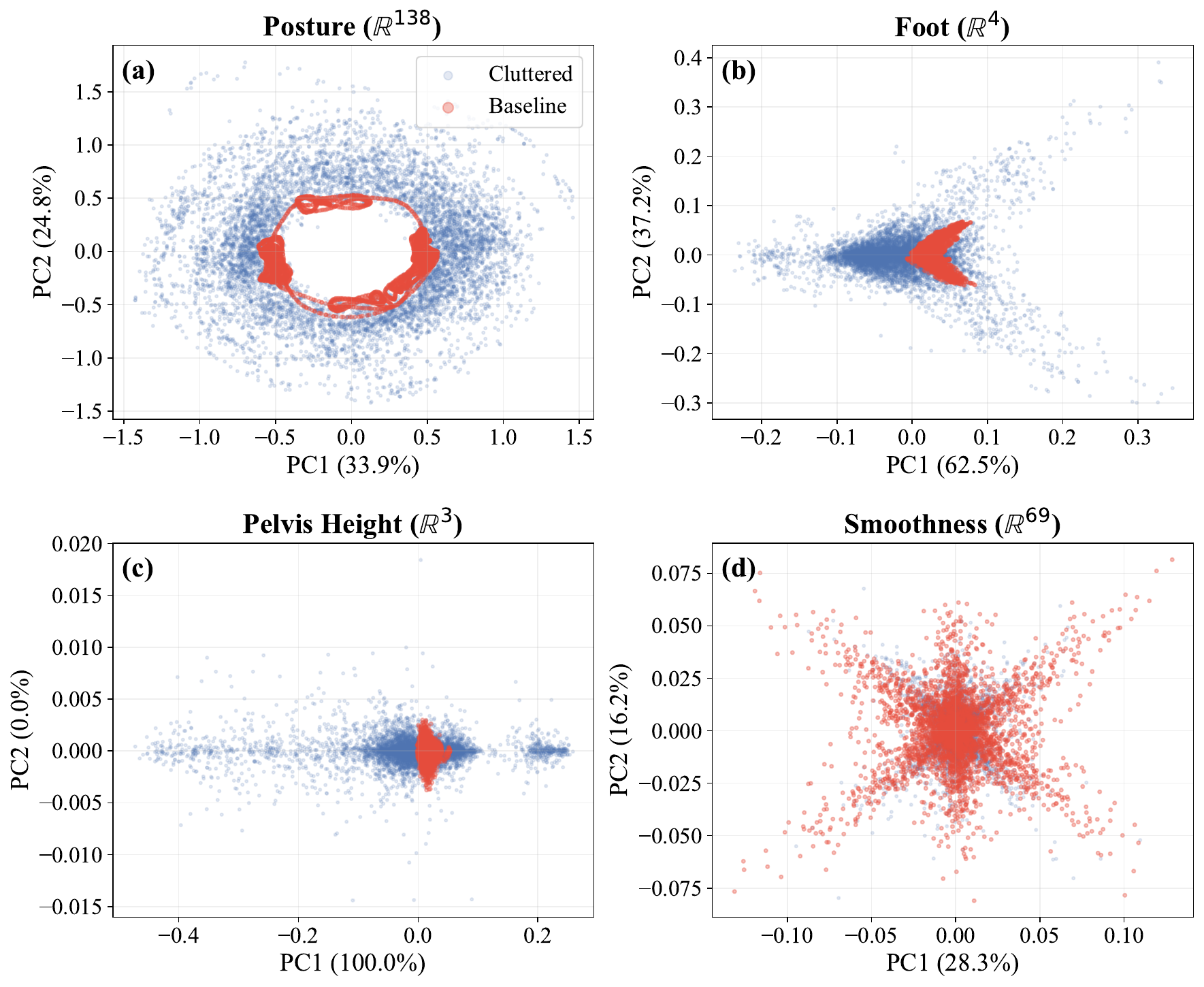}
\vspace{-16pt}
\caption{PCA projections of per-frame kinematic features for four subspaces. \textcolor{baseline_red}{Red}: baseline level-ground walking; \textcolor{clutter_blue}{Blue}: cluttered-environment trajectories.}
\label{fig:fld_pca}
\vspace{-16pt}
\end{figure}

To analyze the behavioral coverage of the collected trajectories, we examine the distribution of per-frame kinematic features across four subspaces, defined by MTC benchmark. For each subspace, feature vectors are extracted from all trajectories as well as from a baseline recording of unobstructed level-ground walking, and projected onto the first two principal components for visualization (Fig.~\ref{fig:fld_pca}).

In the \textit{posture} subspace, baseline frames form a compact closed orbit consistent with a regular gait cycle. 
In the \textit{foot} subspace, the baseline distribution concentrates into a narrow, phase-structured wedge with two symmetric lobes, reflecting alternating left--right foot interactions rather than a closed loop. Similarly, in the \textit{pelvis height} subspace, baseline frames concentrate within a narrow band corresponding to the nearly constant pelvis height of normal walking. 
Across these subspaces, the dataset distributions extend substantially beyond the baseline patterns, indicating diverse posture and foot-interaction adaptations induced by cluttered environments.

In the \textit{smoothness} subspace, the baseline and dataset distributions largely overlap, indicating that the collected trajectories remain consistently smooth despite geometric constraints—a property not always achieved by learned robot policies. 

Together, these projections confirm that the dataset captures a broad spectrum of posture and foot-interaction adaptations while maintaining high motion quality, providing a challenging and diverse benchmark for locomotion in cluttered environments.

\section{Conclusion and Limitations}
\label{sec:conclusion}
We presented MTC, a framework that comprises a data capturer, a dataset, and a benchmark for data-driven, scene-aware humanoid locomotion in cluttered 3D environments. The MTC Capturer leverages immersive VR to capture human locomotion in virtual 3D cluttered environments at scale. The MTC Dataset couples procedurally generated geometric regimes with embodiment-consistent traversal trajectories, while the MTC Benchmark evaluates trajectories along two complementary axes: geometry-induced kinematic adaptation and collision safety under full scene geometry. 

Through dataset-level analysis and goal-conditioned case studies, we demonstrate that MTC captures diverse whole-body adaptation induced by rich spatial constraints, rather than merely diversifying different scene layouts. Quantitative results show that distinct geometric regimes elicit measurably different traversal behaviors from human demonstrators, revealing structured adaptation patterns that existing benchmarks, which vary only scene layout or goal placement, are insufficient to characterize.
Furthermore, the proposed benchmark metrics provide a principled protocol for analyzing locomotion performance under spatial constraints, offering actionable guidance for downstream algorithm development. 

Preliminary results show that a reinforcement learning-based motion tracking policy, trained to imitate MTC trajectories, can reproduce geometry-induced traversal behaviors with low collision rates.

Despite these contributions, several limitations remain. First, the current pipeline employs scene-agnostic retargeting; fully scene-aware motion generation remains an open challenge requiring advances in control and learning. Second, scene layout generation relies on manually designed placement priors rather than learned distributions, which may not fully capture real-world variability. Integrating generative scene models such as Vision-Language Models could improve realism and diversity. Third, the dataset focuses on locomotion-centric traversal and does not model contact-assisted progression, which may be necessary in extreme clutter where multi-contact support is required for balance. Finally, the VR-based motion capture relies on pose estimation and may introduce tracking noise. Our future work will incorporate high-precision motion capture to further improve accuracy.

In summary, MTC provides a foundation for systematic study of geometry-induced adaptation in humanoid locomotion and stimulates further research on scene-aware humanoid planning and control with scalable data collection.

\bibliographystyle{IEEEtran}
\bibliography{IEEEabrv}

@article{radosavovic2024humanoid,
  author    = {Radosavovic, Ilija and Xiao, Tete and Zhang, Bike and Darrell, Trevor and Malik, Jitendra and Sreenath, Koushil},
  title     = {Real-World Humanoid Locomotion with Reinforcement Learning},
  journal   = {Sci. Robot.},
  volume    = {9},
  number    = {89},
  pages     = {eadi9579},
  year      = {2024},
  doi       = {10.1126/scirobotics.adi9579}
}

@article{haarnoja2024soccer,
  author    = {Haarnoja, Tuomas and Moran, Ben and Lever, Guy and Huang, Sandy H. and Tirumala, Dhruva and Humplik, Jan and Wulfmeier, Markus and Tunyasuvunakool, Saran and Siegel, Noah Y. and Hafner, Roland and others},
  title     = {Learning Agile Soccer Skills for a Bipedal Robot with Deep Reinforcement Learning},
  journal   = {Sci. Robot.},
  volume    = {9},
  number    = {89},
  pages     = {eadi8022},
  year      = {2024},
  doi       = {10.1126/scirobotics.adi8022}
}

@article{peng2021amp,
  author    = {Peng, Xue Bin and Ma, Ze and Abbeel, Pieter and Levine, Sergey and Kanazawa, Angjoo},
  title     = {{AMP}: Adversarial Motion Priors for Stylized Physics-Based Character Animation},
  journal   = {ACM Trans. Graph.},
  volume    = {40},
  number    = {4},
  pages     = {1--20},
  year      = {2021},
  doi       = {10.1145/3450626.3459670}
}

@article{peng2018deepmimic,
  author    = {Peng, Xue Bin and Abbeel, Pieter and Levine, Sergey and van de Panne, Michiel},
  title     = {{DeepMimic}: Example-Guided Deep Reinforcement Learning of Physics-Based Character Skills},
  journal   = {ACM Trans. Graph.},
  volume    = {37},
  number    = {4},
  pages     = {1--14},
  year      = {2018},
  doi       = {10.1145/3197517.3201311}
}

@inproceedings{cheng2024exbody,
  author    = {Cheng, Xuxin and Ji, Yandong and Chen, Junming and Yang, Ruihan and Yang, Ge and Wang, Xiaolong},
  title     = {Expressive Whole-Body Control for Humanoid Robots},
  booktitle = {Proc. Robot., Sci. Syst. ({RSS})},
  year      = {2024}
}

@article{ji2024exbody2,
  author    = {Ji, Mazeyu and Peng, Xuanbin and Liu, Fangchen and Li, Jialong and Yang, Ge and Cheng, Xuxin and Wang, Xiaolong},
  title     = {{ExBody2}: Advanced Expressive Humanoid Whole-Body Control},
  journal   = {arXiv preprint},
  eprint    = {2412.13196},
  year      = {2024}
}

@article{xue2025humanoidpf,
  author    = {Xue, Han and Liang, Sikai and Zhang, Zhikai and Zeng, Zicheng and Liu, Yun and Lian, Yunrui and Wang, Jilong and Liu, Qingtao and Shi, Xuesong and Yi, Li},
  title     = {Collision-Free Humanoid Traversal in Cluttered Indoor Scenes},
  journal   = {arXiv preprint},
  eprint    = {2601.16035},
  year      = {2025}
}

@inproceedings{mahmood2019amass,
  author    = {Mahmood, Naureen and Ghorbani, Nima and Troje, Nikolaus F. and Pons-Moll, Gerard and Black, Michael J.},
  title     = {{AMASS}: Archive of Motion Capture as Surface Shapes},
  booktitle = {Proc. IEEE/CVF Int. Conf. Comput. Vis. ({ICCV})},
  pages     = {5442--5451},
  year      = {2019}
}

@inproceedings{guo2022humanml3d,
  author    = {Guo, Chuan and Zou, Shihao and Zuo, Xinxin and Wang, Sen and Ji, Wei and Li, Xingyu and Cheng, Li},
  title     = {Generating Diverse and Natural {3D} Human Motions from Text},
  booktitle = {Proc. IEEE/CVF Conf. Comput. Vis. Pattern Recognit. ({CVPR})},
  pages     = {5152--5161},
  year      = {2022}
}

@inproceedings{lin2023motionx,
  author    = {Lin, Jing and Zeng, Ailing and Lu, Shunlin and Cai, Yuanhao and Zhang, Ruimao and Wang, Haoqian and Zhang, Lei},
  title     = {{Motion-X}: A Large-Scale {3D} Expressive Whole-Body Human Motion Dataset},
  booktitle = {Proc. Adv. Neural Inf. Process. Syst. ({NeurIPS})},
  year      = {2023}
}

@inproceedings{punnakkal2021babel,
  author    = {Punnakkal, Abhinanda R. and Chandrasekaran, Arjun and Athanasiou, Nikos and Quiros-Ramirez, Alejandra and Black, Michael J.},
  title     = {{BABEL}: Bodies, Action and Behavior with {E}nglish Labels},
  booktitle = {Proc. IEEE/CVF Conf. Comput. Vis. Pattern Recognit. ({CVPR})},
  pages     = {722--731},
  year      = {2021}
}

@inproceedings{hassan2019prox,
  author    = {Hassan, Mohamed and Choutas, Vasileios and Tzionas, Dimitrios and Black, Michael J.},
  title     = {Resolving {3D} Human Pose Ambiguities with {3D} Scene Constraints},
  booktitle = {Proc. IEEE/CVF Int. Conf. Comput. Vis. ({ICCV})},
  pages     = {2282--2292},
  year      = {2019}
}

@inproceedings{hassan2021samp,
  author    = {Hassan, Mohamed and Ceylan, Duygu and Villegas, Ruben and Saito, Jun and Yang, Jimei and Zhou, Yi and Black, Michael J.},
  title     = {Stochastic Scene-Aware Motion Prediction},
  booktitle = {Proc. IEEE/CVF Int. Conf. Comput. Vis. ({ICCV})},
  pages     = {11374--11384},
  year      = {2021}
}

@inproceedings{wang2022humanise,
  author    = {Wang, Zan and Chen, Yixin and Liu, Tengyu and Zhu, Yixin and Liang, Wei and Huang, Siyuan},
  title     = {{HUMANISE}: Language-Conditioned Human Motion Generation in {3D} Scenes},
  booktitle = {Proc. Adv. Neural Inf. Process. Syst. ({NeurIPS})},
  year      = {2022}
}

@inproceedings{jiang2024trumans,
  author    = {Jiang, Nan and Zhang, Zhiyuan and Li, Hongjie and Ma, Xiaoxuan and Wang, Zan and Chen, Yixin and Liu, Tengyu and Zhu, Yixin and Huang, Siyuan},
  title     = {{TRUMANS}: Scaling Up Dynamic Human-Scene Interaction Modeling},
  booktitle = {Proc. IEEE/CVF Conf. Comput. Vis. Pattern Recognit. ({CVPR})},
  year      = {2024}
}

@inproceedings{fu2024humanplus,
  author    = {Fu, Zipeng and Zhao, Qingqing and Wu, Qi and Wetzstein, Gordon and Finn, Chelsea},
  title     = {{HumanPlus}: Humanoid Shadowing and Imitation from Humans},
  booktitle = {Proc. Conf. Robot Learn. ({CoRL})},
  year      = {2024}
}

@inproceedings{he2024h2o,
  author    = {He, Tairan and Luo, Zhengyi and Xiao, Wenli and Zhang, Chong and Kitani, Kris and Liu, Changliu and Shi, Guanya},
  title     = {Learning Human-to-Humanoid Real-Time Whole-Body Teleoperation},
  booktitle = {Proc. IEEE/RSJ Int. Conf. Intell. Robots Syst. ({IROS})},
  year      = {2024}
}

@inproceedings{he2024omnih2o,
  author    = {He, Tairan and Luo, Zhengyi and He, Xialin and Xiao, Wenli and Zhang, Chong and Zhang, Weinan and Kitani, Kris and Liu, Changliu and Shi, Guanya},
  title     = {{OmniH2O}: Universal and Dexterous Human-to-Humanoid Whole-Body Teleoperation and Learning},
  booktitle = {Proc. Conf. Robot Learn. ({CoRL})},
  year      = {2024}
}

@inproceedings{cheng2024opentv,
  author    = {Cheng, Xuxin and Li, Jialong and Yang, Shiqi and Yang, Ge and Wang, Xiaolong},
  title     = {{Open-TeleVision}: Teleoperation with Immersive Active Visual Feedback},
  booktitle = {Proc. Conf. Robot Learn. ({CoRL})},
  year      = {2024}
}

@article{luo2025sonic,
  author    = {Luo, Zhengyi and Yuan, Ye and Wang, Tingwu and Li, Chenran and Chen, Sirui and Castaneda, Fernando and Cao, Zi-Ang and Li, Jiefeng and others},
  title     = {{SONIC}: Supersizing Motion Tracking for Natural Humanoid Whole-Body Control},
  journal   = {arXiv preprint},
  eprint    = {2511.07820},
  year      = {2025}
}

@article{ze2025twist,
  author    = {Ze, Yanjie and Chen, Zixuan and Araujo, Joao Pedro and Cao, Zi-ang and Peng, Xue Bin and Wu, Jiajun and Liu, C. Karen},
  title     = {{TWIST}: Teleoperated Whole-Body Imitation System},
  journal   = {arXiv preprint},
  eprint    = {2505.02833},
  year      = {2025}
}

@article{ze2025twist2,
  author    = {Ze, Yanjie and Zhao, Siheng and Wang, Weizhuo and Kanazawa, Angjoo and Duan, Rocky and Abbeel, Pieter and Shi, Guanya and Wu, Jiajun and Liu, C. Karen},
  title     = {{TWIST2}: Scalable, Portable, and Holistic Humanoid Data Collection System},
  journal   = {arXiv preprint},
  eprint    = {2511.02832},
  year      = {2025}
}

@inproceedings{araujo2026gmr,
  author    = {Araujo, Joao Pedro and Ze, Yanjie and Xu, Pei and Wu, Jiajun and Liu, C. Karen},
  title     = {Retargeting Matters: General Motion Retargeting for Humanoid Motion Tracking},
  booktitle = {Proc. IEEE Int. Conf. Robot. Autom. ({ICRA})},
  year      = {2026}
}

@article{yang2025omniretarget,
  author    = {Yang, Lujie and Huang, Xiaoyu and Wu, Zhen and Kanazawa, Angjoo and Abbeel, Pieter and Sferrazza, Carmelo and Liu, C. Karen and Duan, Rocky and Shi, Guanya},
  title     = {{OmniRetarget}: Interaction-Preserving Data Generation for Humanoid Whole-Body Loco-Manipulation and Scene Interaction},
  journal   = {arXiv preprint},
  eprint    = {2509.26633},
  year      = {2025}
}

@inproceedings{deitke2022procthor,
  author    = {Deitke, Matt and VanderBilt, Eli and Herrasti, Alvaro and Weihs, Luca and Salvador, Jordi and Ehsani, Kiana and Han, Winson and Kolve, Eric and Farhadi, Ali and Kembhavi, Aniruddha and Mottaghi, Roozbeh},
  title     = {{ProcTHOR}: Large-Scale Embodied {AI} Using Procedural Generation},
  booktitle = {Proc. Adv. Neural Inf. Process. Syst. ({NeurIPS})},
  year      = {2022}
}

@inproceedings{raistrick2023infinigen,
  author    = {Raistrick, Alexander and Lipson, Lahav and Ma, Zeyu and Mei, Lingjie and Wang, Mingzhe and Zuo, Yiming and Kayan, Karhan and Wen, Hongyu and Han, Beining and Wang, Yihan and Newell, Alejandro and Law, Hei and Goyal, Ankit and Yang, Kaiyu and Deng, Jia},
  title     = {Infinite Photorealistic Worlds Using Procedural Generation},
  booktitle = {Proc. IEEE/CVF Conf. Comput. Vis. Pattern Recognit. ({CVPR})},
  pages     = {12630--12641},
  year      = {2023}
}

@inproceedings{raistrick2024infinigen_indoors,
  author    = {Raistrick, Alexander and Mei, Lingjie and Kayan, Karhan and Yan, David and Zuo, Yiming and Han, Beining and Wen, Hongyu and Parakh, Meenal and Alexandropoulos, Stamatis and Lipson, Lahav and Ma, Zeyu and Deng, Jia},
  title     = {Infinigen Indoors: Photorealistic Indoor Scenes Using Procedural Generation},
  booktitle = {Proc. IEEE/CVF Conf. Comput. Vis. Pattern Recognit. ({CVPR})},
  year      = {2024}
}

@inproceedings{savva2019habitat,
  author    = {Savva, Manolis and Kadian, Abhishek and Maksymets, Oleksandr and Zhao, Yili and Wijmans, Erik and Jain, Bhavana and Straub, Julian and Liu, Jia and Koltun, Vladlen and Malik, Jitendra and Parikh, Devi and Batra, Dhruv},
  title     = {Habitat: A Platform for Embodied {AI} Research},
  booktitle = {Proc. IEEE/CVF Int. Conf. Comput. Vis. ({ICCV})},
  pages     = {9339--9347},
  year      = {2019}
}

@inproceedings{yang2024holodeck,
  author    = {Yang, Yue and Fan, Fan-Yun and Dickinson, Kaizhi and Wu, Jialu and Khashabi, Daniel and Choi, Yejin and Kembhavi, Aniruddha},
  title     = {Holodeck: Language Guided Generation of {3D} Embodied {AI} Environments},
  booktitle = {Proc. IEEE/CVF Conf. Comput. Vis. Pattern Recognit. ({CVPR})},
  year      = {2024}
}

@article{kim2024acrobatic,
  author    = {Kim, Dohyeong and Bae, Junwon and Lee, Junhyeok and Son, Dongjoon and Oh, Songhwai},
  title     = {Stage-Wise Reward Shaping for Acrobatic Robots: A Constrained Multi-Objective Reinforcement Learning Approach},
  journal   = {arXiv preprint},
  eprint    = {2409.15755},
  year      = {2024}
}

@article{wang2026omnixtreme,
  title     = {{OmniXtreme}: Breaking the Generality Barrier in High-Dynamic Humanoid Control},
  author    = {Wang, Yunshen and Zhu, Shaohang and Zhi, Peiyuan and Li, Yuhan and Li, Jiaxin and Li, Yong-Lu and Xiao, Yuchen and Wang, Xingxing and Jia, Baoxiong and Huang, Siyuan},
  journal   = {arXiv preprint},
  eprint    = {2602.23843},
  year      = {2026}
}

\end{document}